\documentclass[conference]{IEEEtran}
\usepackage{algorithm}
\usepackage{algorithmic}
\IEEEoverridecommandlockouts
\usepackage{amsmath}
\usepackage{amssymb}
\usepackage{mathtools}
\usepackage{amsthm}
\usepackage{verbatim}
\DeclareMathOperator{\Log}{Log}
\DeclareMathOperator{\Exp}{Exp}

\usepackage{graphicx}
\usepackage{subcaption}

\theoremstyle{plain}
\newtheorem{theorem}{Theorem}[section]

\newtheorem{lemma}[theorem]{Lemma}

\theoremstyle{definition}

\newtheorem{assumption}[theorem]{Assumption}
\theoremstyle{remark}
\newtheorem{remark}[theorem]{Remark}
\usepackage{cite}
\usepackage{amsmath,amssymb,amsfonts}
\usepackage{algorithmic}
\usepackage{graphicx}
\usepackage{textcomp}
\usepackage{xcolor}
\usepackage{booktabs}
\usepackage{url} 

\def\BibTeX{{\rm B\kern-.05em{\sc i\kern-.025em b}\kern-.08em
    T\kern-.1667em\lower.7ex\hbox{E}\kern-.125emX}}
\begin{document}

\title{A2G-QFL: Adaptive Aggregation with Two Gains in Quantum Federated learning 
\\
}

\author{\IEEEauthorblockN{Shanika Iroshi Nanayakkara}
\IEEEauthorblockA{\textit{School of IT},
\textit{Deakin University}\\
Geelong, Australia \\
s.nanayakkara@deakin.edu.au}
\and
\IEEEauthorblockN{Shiva Raj Pokhrel}
\IEEEauthorblockA{\textit{School of IT},
\textit{Deakin University}\\
Geelong, Australia \\
shiva.pokhrel@deakin.edu.au}
}

\maketitle

\begin{abstract}
Federated learning (FL) deployed over quantum-enabled and heterogeneous classical
networks face significant performance degradation due to uneven client quality,
stochastic teleportation fidelity, device instability, and geometric mismatch between
local and global models. Classical aggregation rules assume euclidean topology and
uniform communication reliability, limiting their suitability for emerging quantum
federated systems. This paper introduces \textbf{A2G (Adaptive Aggregation with Two
Gains)}, a dual-gain framework that jointly regulates geometric blending through a
geometry gain and modulates client importance using a QoS gain derived from
teleportation fidelity, latency, and instability. We develop the A2G update rule,
establish convergence guarantees under smoothness and bounded-variance assumptions,
and show that A2G recovers FedAvg, QoS-aware averaging, and manifold-based
aggregation as special cases. Experiments on a quantum--classical hybrid testbed
demonstrate improved stability and higher accuracy under heterogeneous and noisy
conditions. 

\end{abstract}


\section{Introduction}

Federated learning (FL)~\cite{mcmahan2017communication, nanayakkara2024understanding} deployed over quantum-enabled and heterogeneous classical networks exhibits several fundamental challenges. These include non-uniform client reliability, stochastic variations in teleportation fidelity, device-level instability, circular or manifold-valued model parameters (e.g., rotation angles in variational quantum circuits), and geometry-induced drift among local model trajectories.

Conventional aggregation schemes implicitly assume a Euclidean parameter space and homogeneous client behaviour, assumptions that break down in quantum-assisted federated settings. To address these limitations, we introduce \textbf{A2G (Adaptive Aggregation with Two Gains)}, a dual-gain aggregation framework designed to incorporate both client quality and model geometry into the global update process.

A2G  as shown in Fig.~1 operates by (i) applying a \emph{QoS gain}~$\alpha$ that adaptively modulates client trust based on teleportation fidelity, communication latency, and local model stability, and (ii) employing a \emph{geometry gain}~$\beta$ that adjusts global updates according to the curvature and non-Euclidean structure of the underlying parameter space. Existing approaches—including FedAvg, FedProx, and Riemannian or manifold-aware averaging—fail to jointly resolve both Quality-of-Service (QoS) heterogeneity and geometric nonlinearity, motivating the need for a multi-gain adaptive mechanism.


\begin{figure}[t]
     \centering
    \includegraphics[width=1\linewidth]{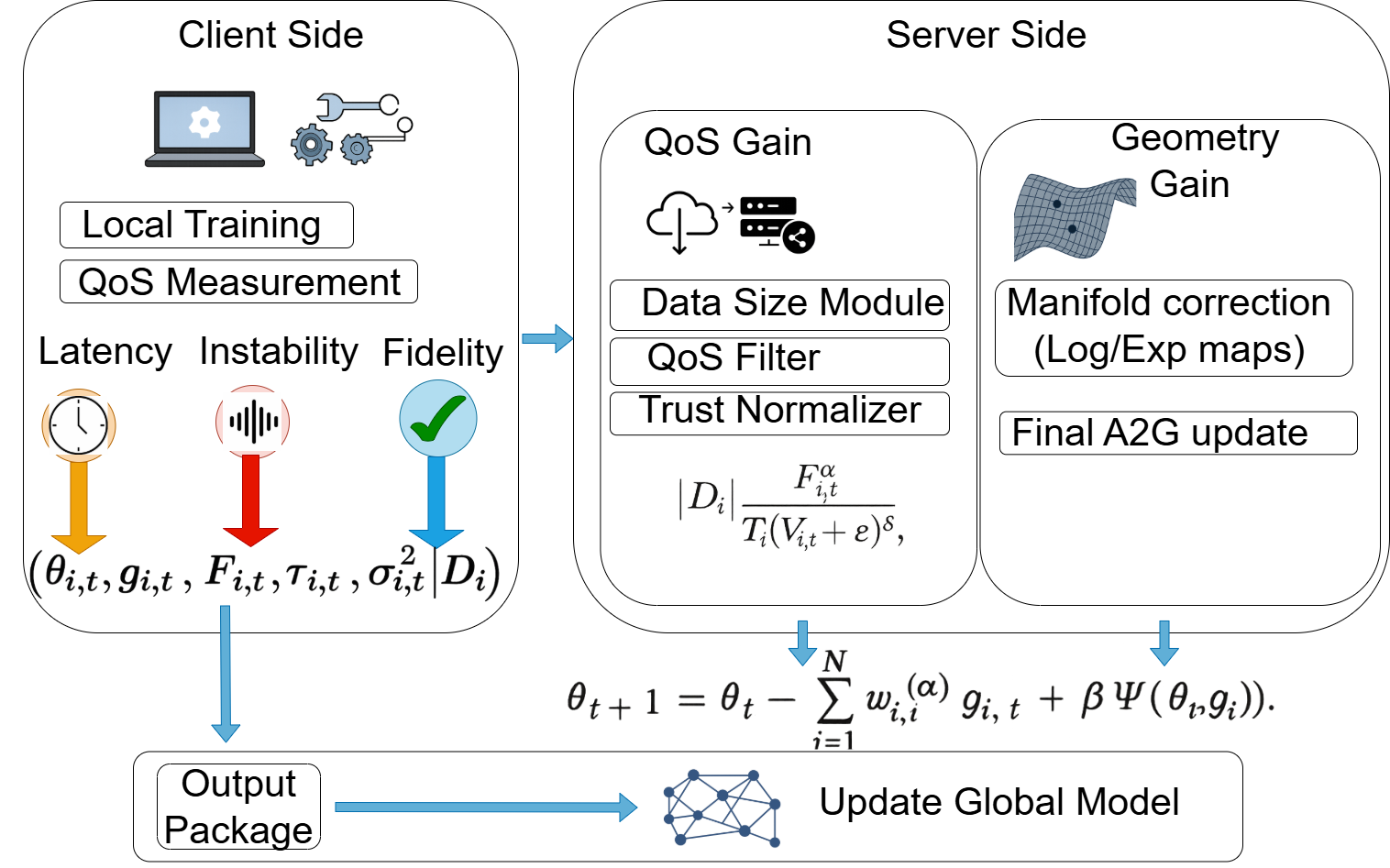}
  \caption{Quantum Federated Learning: A High Level View of Adaptive Aggregation with Two Gains}
\label{fig} 
\end{figure}

\subsection{Limitations of Prior Work}
FedAvg does not account for geometric considerations or network QoS. Riemannian FL lacks integration of latency and fidelity metrics. Quantum FL research primarily addresses quantum model development rather than aggregation strategies \cite{liu2025practical}.
Federated learning has traditionally relied on euclidean parameter averaging, as exemplified by FedAvg. However, FedAvg assumes euclidean averaging and demonstrates poor performance in non-independent and identically distributed (non-IID) settings.
This approach, as described by McMahan et al. \cite{mcmahan2017communication}, implicitly assumes that local model updates occur within a locally linear region of the loss landscape.
Extensive empirical and theoretical analyses indicate that statistical heterogeneity, device variability, and client drift result in highly non-linear and curved parameter trajectories. These factors render Euclidean averaging unstable and frequently divergent in heterogeneous environments
\cite{karimireddy2020scaffold,li2020federated,wang2020tackling}.
This development has led to an increasing body of research on Riemannian and geometry-aware federated optimization, in which model parameters are represented as points on smooth manifolds and aggregation is conducted using exponential and logarithmic maps \cite{bonnabel2013stochastic, sakai2021riemannian, zhang2016riemannian, hsieh2023riemannian, zhang2024nonconvex}. 

The geometric challenges are even more pronounced in quantum federated learning (QFL), where variational quantum circuits parameterize unitary transformations via angles living on periodic manifolds \cite{mitarai2018quantum,schuld2019quantum,cerezo2021variational,abbas2021power, khan2025dt}. These parameters evolve under the influence of hardware noise, decoherence, shot-induced gradient variance, and in distributed implementations teleportation fidelity fluctuations, all of which distort the effective optimization landscape and make naive euclidean aggregation misaligned with the underlying parameter geometry \cite{beer2020training,huang2025vast}. Prior QFL studies either disregard geometric structure or address it heuristically, resulting in unstable global updates and high variance under quantum channel imperfections \cite{chen2021federated, qi2023optimizing}.
However, A2G provides a general dual-gain aggregation framework for mixed classical–quantum FL, whereas FAAA-$\beta$ which we proposed in our previous work is a specialized teleportation-conditioned angular consensus rule for variational quantum models operating on toroidal parameter spaces.

Consequently, a principled geometry-control mechanism---captured through a geometry gain $\beta$ and a manifold correction operator $\Psi$---is required in both classical and quantum FL to ensure that aggregation respects intrinsic curvature, damps divergence, and stabilizes global convergence under statistical and hardware heterogeneity. The A2G framework introduced in this work unifies these insights by integrating Euclidean, QoS-weighted, and manifold-aware aggregation within a single adaptive dual-gain rule, thereby providing a universal mechanism applicable to classical, quantum, and hybrid federated learning systems.

\begin{table}[t]
\centering
\caption{Notation Used in A2G-QFL}
\begin{tabular}{ll}
\toprule
\textbf{Symbol} & \textbf{Meaning} \\
\midrule
$K$ & Number of clients in the federated system \\
$\theta_t$ & Global model at communication round $t$ \\
$\theta_{i,t}$ & Local model of client $i$ at round $t$ \\
$g_{i,t}$ & Local (stochastic) gradient or SPSA estimate \\
$D_i$ & Dataset held by client $i$ \\
$|D_i|$ & Size of client $i$'s dataset \\
$p_i$ & Data-size proportion: $p_i = |D_i| / \sum_j |D_j|$ \\
$f_i(\theta)$ & Local loss function of client $i$ \\
$f(\theta)$ & Global loss: $f(\theta)=\sum_{i=1}^K p_i f_i(\theta)$ \\
\midrule
$F_{i,t}$ & Teleportation fidelity (link quality) of client $i$ at round $t$ \\
$\tau_{i,t}$ & Communication latency for client $i$ \\
$\sigma^2_{i,t}$ & Instability / variance of client $i$ \\
$q_{i,t}$ & QoS factor: $F_{i,t}^{\alpha} (\tau_{i,t}+\varepsilon)^{-\gamma}(\sigma_{i,t}^2+\varepsilon)^{-\delta}$ \\
$\hat{w}_{i,t}^{(\alpha)}$ & QoS-weighted trust score: $p_i q_{i,t}$ \\
$w_{i,t}^{(\alpha)}$ & Normalized QoS trust weight \\
$\alpha$ & Fidelity-sensitive QoS gain exponent \\
$\gamma,\delta$ & Latency and instability gain exponents \\
\midrule
$\mathcal{M}$ & Parameter manifold (e.g., $\mathbb{R}^d$, $S^1$, $T^d$) \\
$\Log_{\theta_t}(\cdot)$ & Riemannian logarithm map at $\theta_t$ \\
$\Exp_{\theta_t}(\cdot)$ & Riemannian exponential map at $\theta_t$ \\
$v_t$ & QoS-weighted tangent vector: $\sum_i w_{i,t}^{(\alpha)} \Log_{\theta_t}(\theta_{i,t})$ \\
$\Psi(\theta_t,\{\theta_{i,t}\})$ & Geometry correction: $\Exp_{\theta_t}(v_t) - \theta_t$ \\
$\beta$ & Geometry gain controlling curvature-aware blending \\
$\rho_t$ & Client deviation radius: $\max_i \|\theta_{i,t} - \theta_t\|$ \\
$\rho$ & Worst-case dispersion: $\sup_t \rho_t$ \\
\midrule
$\eta_t$ & Server-side learning rate (set to $0$ in experiments) \\
$\varepsilon$ & Small regularizer preventing division by zero \\
$L$ & Smoothness constant of $f_i$ \\
$\sigma^2$ & Bounded gradient variance \\
$G$ & Second-moment bound on gradients \\
\bottomrule
\end{tabular}
\end{table}

\subsection{Key Contributions}

Our key contributions are as follows:

\begin{enumerate}
    \item \textit{A universal dual-gain aggregation framework.}
    We introduce A2G-QFL, the first FL aggregation mechanism that jointly incorporates:
    (i) a \emph{QoS gain}~$\alpha$ based on teleportation fidelity, latency, and instability; and
    (ii) a \emph{geometry gain}~$\beta$ that regulates curvature-aware blending between local and global models.

    \item \textit{Unified formulation covering classical, quantum, and hybrid FL.}
    A2G recovers FedAvg, QoS-weighted averaging, and manifold-based aggregation as special cases.
    This unifies Euclidean, quantum, and Riemannian/topological FL under a single formulation that is immediately deployable across classical and quantum clients.

    \item \textit{Convergence guarantees under heterogeneous QoS.}
    We derive a complete convergence analysis for A2G under smoothness, bounded-variance, and finite-QoS assumptions.
    The resulting bound explicitly reveals how QoS gains accelerate convergence, while geometry gains introduce curvature-aware correction.

    \item \textit{Teleportation-aware trust weighting.}
    We formalize a QoS-driven trust mechanism where client reliability is modulated by teleportation fidelity, communication delay, and model instability—providing the first reliability-aware aggregation scheme tailored to noisy quantum channels.

    \item \textit{Geometry-controlled stabilization of noisy QNN updates.}
    We show that small geometry gains ($\beta \ll 1$) suppress client drift and curvature-induced bias in quantum models, significantly improving global model stability compared to traditional averaging.

    \item \textit{Robust empirical improvements under noise and heterogeneity.}
    Experiments on a quantum--classical hybrid testbed demonstrate that A2G yields consistently higher accuracy and stability under heterogeneous and noisy conditions.
    For example, $\beta = 0.05$ achieves $68.25\%$ best accuracy $(13.65\%$ or $ {\approx 25\% }$ relative improvement over $\beta=1.0$), and this improvement persists even under high teleportation noise.

    \item \textit{Generalizable mechanism for next-generation distributed quantum intelligence.}
    A2G provides a communication- and geometry-aware aggregation rule that is compatible with classical FL, quantum FL, and hybrid deployments, enabling scalable and noise-resilient federated learning in emerging quantum-assisted networks.
\end{enumerate}

\section{Methodology-A2G: Adaptive Aggregation with Two Gains}
\label{sec:method}

We develop a universal aggregation mechanism (A2G) for heterogeneous federated
learning operating across quantum-enabled and classical communication
environments. The framework jointly incorporates (i) communication quality
through QoS-sensitive trust weighting and (ii) model geometry through a
controllable interpolation between Euclidean and Riemannian averaging.
This section formalizes the methodological construction, organized around the
four components implemented in Algorithms~\ref{alg:a2g_main}--\ref{alg:geo_a2g}.

\subsection{Problem Setting and Notation}

Let $\theta_t \in \mathcal{M}$ denote the global model at communication round $t$,
where $\mathcal{M}$ is either the Euclidean space $\mathbb{R}^d$ or a
Riemannian manifold endowed with metric~$g$.  
Client~$i$ holds dataset $D_i$ with cardinality $|D_i|$
and local loss function $f_i(\theta)$.
The global loss is
\begin{equation}
    f(\theta) = \sum_{i=1}^K p_i f_i(\theta),
    \qquad
    p_i = \frac{|D_i|}{\sum_{j=1}^K |D_j|}.
\end{equation}

\begin{algorithm}[h]
\caption{A2G: Adaptive Aggregation with QoS Gain $\alpha, \gamma, \delta$ and Geometry Gain $\beta$}
\label{alg:a2g_main}
\begin{algorithmic}[1]
  \REQUIRE Current global parameter $\theta_t$; client datasets $\{D_i\}_{i=1}^K$;
           QoS gains $(\alpha,\gamma,\delta)$; geometry gain $\beta$; step size $\eta$;
           (optional) manifold operators $\Log_{\theta_t}(\cdot)$, $\Exp_{\theta_t}(\cdot)$.
  \ENSURE Updated global parameter $\theta_{t+1}$.

  \STATE \textbf{Phase I: Client-side local updates and QoS reporting}
  \FORALL{$i \in \{1,\dots,K\}$ \textbf{in parallel}}
    \STATE $(\theta_{i,t}, g_{i,t}, F_{i,t}, \tau_{i,t}, \sigma_{i,t}^2, |D_i|);\text{ClientUpdate}(i, \theta_t, D_i)$
  \ENDFOR

  \STATE \textbf{Phase II: QoS weighting via $\alpha$-gain}
  \STATE $\textnormal{QoS}_t := \{(F_{i,t}, \tau_{i,t}, \sigma_{i,t}^2)\}_{i=1}^K.$
  \STATE $\{w_{i,t}^{(\alpha)}\}_{i=1}^K \gets
    \textsc{QoSWeights}(\text{QoS}_t, \{|D_i|\};\,\alpha,\gamma,\delta)$
 
  \STATE \textbf{Phase III: Geometry-controlled A2G update}
\STATE $\theta_{t+1} \gets
 \textsc{GeoA2GUpdate}(\ldots)$ 
 \COMMENT{QoS-weighted gradient term $(-\eta g_t^{\mathrm{agg}})$ + geometry gain $\beta \Psi(\cdot)$}

  \RETURN $\theta_{t+1}$
\end{algorithmic}
\end{algorithm}

In the QoS gain $\alpha$ controls the sensitivity of the aggregation weights $w_{i,t}^{(\alpha)}$ 
to teleportation fidelity, latency, and instability, as formalized in Algorithm~\ref{alg:qosweights}. 
The \emph{geometry gain} $\beta$ modulates the contribution of the manifold correction 
$\Psi(\theta_t,\{\theta_{i,t}\})$ in Algorithm~\ref{alg:geo_a2g}, thereby interpolating between 
purely Euclidean aggregation ($\beta = 0$) and fully geometry-aware averaging ($\beta = 1$) on 
the underlying parameter manifold (e.g., toroidal, circular, or more general Riemannian geometry). 
Together, $(\alpha,\beta)$ specify a family of QoS- and geometry-aware federated updates that 
specialize to FedAvg, QoS-weighted FedAvg, and Riemannian averaging as limiting cases.

In each round, clients compute local updates and measure communication quality,
while the server performs QoS-aware and geometry-aware aggregation.  
A2G thus fuses information from the \emph{learning layer}
(model updates, gradients) and the \emph{communication layer}
(teleportation fidelity, latency, instability).

\subsection{Phase I: Client-Side Local Updates and QoS Sensing}

Algorithm~\ref{alg:clientupdate} formalizes the client-side procedure.
Upon receiving $\theta_t$, each client executes a local optimizer (e.g., SPSA)
to obtain a local model $\theta_{i,t}$ and optionally a stochastic gradient
estimate $g_{i,t}$.  
Simultaneously, the client measures three physical-layer QoS indicators:
\begin{itemize}
    \item fidelity $F_{i,t}$ (quantum teleportation or link quality),
    \item classical or hybrid latency $\tau_{i,t}$,
    \item channel instability variance $\sigma^2_{i,t}$.
\end{itemize}

Each client returns the tuple $ (\theta_{i,t},\, g_{i,t},\, F_{i,t},\, \tau_{i,t},\, \sigma_{i,t}^2,\, |D_i|)$
to the server, supplying both model updates and physical-layer reliability
information.
In our experiments, server-side gradients are disabled by setting $\eta=0$;
nonetheless, the full A2G formulation supports $\eta>0$ when gradient-based
server updates are desired.

\begin{algorithm}[h]
\caption{ClientUpdate: Local Training and QoS Measurement on Client $i$}
\label{alg:clientupdate}
\begin{algorithmic}[1]
  \REQUIRE Client index $i$; global parameter $\theta_t$; local dataset $D_i$.
  \ENSURE Local parameter $\theta_{i,t}$, gradient estimate $g_{i,t}$,
          fidelity $F_{i,t}$, latency $\tau_{i,t}$, instability $\sigma_{i,t}^2$,
          and shard size $|D_i|$.

  \STATE Initialize local parameter: $\theta \gets \theta_t.$
  \STATE Local optimization on $D_i$: $\theta_{i,t} \gets \textsc{LocalTrain}(\theta, D_i).$
  \STATE Estimate local gradient (or surrogate, optional in purely geometry–driven A2G):
$g_{i,t} \approx \nabla_\theta \ell(\theta_{i,t}; D_i).$

  \STATE Measure QoS indicators on client $i$:$
    F_{i,t},\;\tau_{i,t},\;\sigma_{i,t}^2.$
  \STATE Compute local shard size:$
    |D_i| \gets |D_i|.$
  \RETURN $(\theta_{i,t}, g_{i,t}, F_{i,t}, \tau_{i,t}, \sigma_{i,t}^2, |D_i|)$.
\end{algorithmic}
\end{algorithm}
In our experiments, we focus on the geometry–driven variant of A2G and set
\(\eta = 0\), so the server does not use \(g_{i,t}\); gradients are only used
locally by the clients' optimisers (e.g., SPSA). The full A2G formulation in
Algorithm~\ref{alg:geo_a2g} remains general and supports nonzero \(\eta\) when
server-side gradients are desired in future work.
\subsection{Phase II: QoS Gain and Trust Weight Computation}

Algorithm~\ref{alg:qosweights} defines the QoS-sensitive trust coefficients
$w_{i,t}^{(\alpha)}$, which determine each client's contribution to aggregation.
We first compute the QoS factor
\begin{equation}
  q_{i,t}=\frac{F_{i,t}^{\alpha}}
       {(\tau_{i,t}+\varepsilon)^{\gamma}(\sigma_{i,t}^2+\varepsilon)^{\delta}},
\end{equation}

where $\alpha$, $\gamma$, and $\delta$ regulate sensitivity to fidelity, latency,
and instability, respectively.
This factor is scaled by the normalized data-size proportion $p_i$ to obtain the
trust score $\hat{w}_{i,t}^{(\alpha)} = p_i \, q_{i,t}.$ Finally, normalization yields
\begin{equation}
  w_{i,t}^{(\alpha)} 
  =
  \frac{\hat{w}_{i,t}^{(\alpha)}}
       {\sum_{j=1}^K \hat{w}_{j,t}^{(\alpha)}}.
\end{equation}

Thus, the QoS gain $\alpha$ acts as a tunable control that filters
unreliable clients and amplifies well-performing, high-fidelity nodes.
When $\alpha=0$, A2G reduces to uniform (FedAvg-like) weighting.

\begin{algorithm}[h]
\caption{QoSWeights: QoS- and Data-Size–Aware Trust Coefficients}
\label{alg:qosweights}
\begin{algorithmic}[1]
  \REQUIRE Shard sizes $\{|D_i|\}_{i=1}^K$; fidelities $\{F_{i,t}\}_{i=1}^K$;
           latencies $\{\tau_{i,t}\}_{i=1}^K$; instabilities $\{\sigma_{i,t}^2\}_{i=1}^K$;
           QoS gains $(\alpha,\gamma,\delta)$; small constant $\varepsilon > 0$.
  \ENSURE Normalized trust weights $\{w_{i,t}^{(\alpha)}\}_{i=1}^K$.

  \STATE Compute data-size fractions:
  \STATE \quad $d_{i,t} \gets
      \displaystyle \frac{|D_i|}{\sum_{j=1}^K |D_j|},
      \quad i = 1,\dots,K$

  \STATE Compute QoS factor for each client:
  \STATE \quad $q_{i,t} \gets
      \displaystyle
      \frac{F_{i,t}^{\alpha}}
           {(\tau_{i,t} + \varepsilon)^{\gamma}
            (\sigma_{i,t}^2 + \varepsilon)^{\delta}},
      \quad i = 1,\dots,K$

  \STATE Form unnormalized trust scores:
  \STATE \quad $\hat{w}_{i,t}^{(\alpha)} \gets d_{i,t} \, q_{i,t},
      \quad i = 1,\dots,K$

  \STATE Normalize to obtain aggregation weights:
  \STATE \quad $w_{i,t}^{(\alpha)} \gets
      \displaystyle
      \frac{\hat{w}_{i,t}^{(\alpha)}}
           {\sum_{j=1}^K \hat{w}_{j,t}^{(\alpha)}},
      \quad i = 1,\dots,K$

  \RETURN $\{w_{i,t}^{(\alpha)}\}_{i=1}^K$.
\end{algorithmic}
\end{algorithm}

This formulation couples physical-layer communication quality with learning-layer
aggregation, ensuring that unreliable or noisy clients contribute less than those 
with stable, high-fidelity channels.

If all clients are treated as having equal data size, i.e.\ $p_i = 1/K$,
and the QoS exponents are tied via $\gamma = \delta = \alpha$, then the
general A2G rule reduces to the simpler QoS-only weighting
\begin{equation}
  w_{i,t}^{(\alpha)}
  =
  \frac{
    F_{i,t}^{\alpha}
    (\tau_{i,t}+\varepsilon)^{-\alpha}
    (\sigma_{i,t}^2+\varepsilon)^{-\alpha}
  }{
    \sum_{j=1}^K
    F_{j,t}^{\alpha}
    (\tau_{j,t}+\varepsilon)^{-\alpha}
    (\sigma_{j,t}^2+\varepsilon)^{-\alpha}
  }.
  \label{eq:weight}
\end{equation}

\subsection{Phase III: Geometry Gain and Manifold Correction}

Local models may reside on non-Euclidean parameter spaces, particularly in
quantum circuits where parameters often evolve on periodic manifolds (e.g.,
tori or circles).  
Algorithm~\ref{alg:geo_a2g} introduces the geometry gain $\beta \in [0,1]$
to systematically account for model geometry.

Let $\Log_{\theta_t}$ and $\Exp_{\theta_t}$ denote the Riemannian logarithmic
and exponential maps.  
We compute the QoS-weighted tangent vector
\begin{equation}
  v_t
  =
  \sum_{i=1}^K
    w_{i,t}^{(\alpha)}
    \Log_{\theta_t}(\theta_{i,t}),
\end{equation}
and define the manifold correction term
\begin{equation}
  \Psi(\theta_t,\{\theta_{i,t}\})
  =
  \Exp_{\theta_t}(v_t) - \theta_t.
\end{equation}

The geometry gain $\beta$ interpolates between Euclidean and Riemannian
aggregation:
\begin{itemize}
    \item $\beta=0$: purely Euclidean aggregation (FedAvg-style), and
    \item $\beta=1$: fully geometry-aware Riemannian averaging.
\end{itemize}
In Euclidean space, the map reduces to
\begin{equation}
\Psi(\theta_t,\{\theta_{i,t}\})
=
\sum_{i=1}^K
    w_{i,t}^{(\alpha)}(\theta_{i,t}-\theta_t),
\end{equation}
revealing that geometry-aware A2G generalizes Euclidean aggregation.

\begin{algorithm}[h]
\caption{GeoA2GUpdate: Geometry-Controlled A2G Update}
\label{alg:geo_a2g}
\begin{algorithmic}[1]
  \REQUIRE Current global parameter $\theta_t$; client parameters $\{\theta_{i,t}\}_{i=1}^K$;
           client gradients $\{g_{i,t}\}_{i=1}^K$; trust weights $\{w_{i,t}^{(\alpha)}\}_{i=1}^K$;
           step size $\eta$; geometry gain $\beta$;
           manifold operators $\Log_{\theta_t}(\cdot)$, $\Exp_{\theta_t}(\cdot)$.
  \ENSURE Updated global parameter $\theta_{t+1}$.

  \STATE \textbf{(a) Geometric correction via manifold averaging}
  \STATE \quad $v_t \gets \displaystyle\sum_{i=1}^K w_{i,t}^{(\alpha)} \Log_{\theta_t}(\theta_{i,t})$
  \STATE \quad $\Psi(\theta_t, \{\theta_{i,t}\}) \gets \Exp_{\theta_t}(v_t) - \theta_t$

  \STATE \textbf{(b) QoS-weighted gradient aggregation}
  \STATE \quad $g_t^{\mathrm{agg}} \gets \displaystyle\sum_{i=1}^K w_{i,t}^{(\alpha)} g_{i,t}$

  \STATE \textbf{(c) A2G global update rule}
  \STATE \quad $\theta_{t+1}
      \gets \theta_t - \eta\, g_t^{\mathrm{agg}}
                 + \beta\, \Psi(\theta_t,\{\theta_{i,t}\})$

  \RETURN $\theta_{t+1}$.
\end{algorithmic}
\end{algorithm}

\subsection{Phase IV: Dual-Gain A2G Server Update}

Algorithm~\ref{alg:a2g_main} integrates all components into the A2G update rule:
\begin{equation}
    \theta_{t+1}
    =
    \theta_t
    -
    \eta_t \sum_{i=1}^K w_{i,t}^{(\alpha)} g_{i,t}
    +
    \beta \, \Psi(\theta_t,\{\theta_{i,t}\}).
\end{equation}

\begin{itemize}
    \item The \emph{QoS gain} $(\alpha,\gamma,\delta)$ modulates the trust
    weights applied to both gradients and model deviations.
    \item The \emph{geometry gain} $\beta$ adjusts the curvature-aware blending
    between local and global models.
\end{itemize}

In our experiments we focus on the geometry-driven case ($\eta_t = 0$),
yielding the practical update
\begin{equation}
    \theta_{t+1}
    =
    \theta_t
    +
    \beta
    \sum_{i=1}^K
        w_{i,t}^{(\alpha)}
        (\theta_{i,t} - \theta_t).
\end{equation}

\subsection{Universality Across Classical, Quantum, and Hybrid FL}

Although motivated by quantum-enabled communication, A2G is model- and
system-agnostic.  
If all clients share identical latency and reliability
($F_{i,t}=1$, $\sigma_{i,t}^2=0$, $\tau_{i,t}=\tau$), then
$q_{i,t}$ becomes constant and QoS weighting disappears, recovering classical
FedAvg.  
If parameters lie in Euclidean space and $\beta=0$, the update becomes
fully classical.  
Thus, A2G provides a universal, cross-layer-compatible aggregation framework
that supports classical, quantum, and hybrid federated deployments.

\section{Convergence Analysis of A2G}
\label{sec:convergence}

We analyse the convergence behaviour of A2G under standard smoothness and
stochasticity assumptions used in federated optimisation, augmented with mild
regularity assumptions on teleportation-based QoS indicators. Unless stated
otherwise, we work in the Euclidean case $\mathcal{M} = \mathbb{R}^d$, and
discuss the Riemannian extension at the end.
We consider $K$ clients, each with local loss
\begin{equation}
  f_i(\theta)
  =
  \mathbb{E}_{\xi \sim D_i}\bigl[\ell(\theta;\,\xi)\bigr],
  \qquad i = 1,\dots,K,
\end{equation}
and global objective
\begin{equation}
  f(\theta)
  =
  \sum_{i=1}^K p_i f_i(\theta),
  \qquad
  p_i = \frac{|D_i|}{\sum_{j=1}^K |D_j|}.
\end{equation}

\subsection{Assumptions}

\begin{assumption}[Smoothness]
\label{asm:smooth}
Each local objective $f_i$ is $L$-smooth:
\begin{equation}
  \|\nabla f_i(\theta) - \nabla f_i(\theta')\|
  \;\le\;
  L \|\theta - \theta'\|,
  \quad \forall\,\theta,\theta' \in \mathbb{R}^d.
\end{equation}
Thus the global loss $
  f(\theta) = \sum_{i=1}^K p_i f_i(\theta)$
is also $L$-smooth.
\end{assumption}

\begin{assumption}[Unbiased Stochastic Gradients]
\label{asm:unbiased}
Client stochastic gradients satisfy
\begin{equation}
  \mathbb{E}[g_{i,t} \mid \theta_t] = \nabla f_i(\theta_t), 
  \qquad
  \mathbb{E}\bigl\|g_{i,t} - \nabla f_i(\theta_t)\bigr\|^2 \le \sigma^2,
\end{equation}
for all clients $i$ and rounds $t$.
\end{assumption}

\begin{assumption}[Bounded Second Moment]
\label{asm:secondmoment}
The second moment of each stochastic gradient is bounded:
\begin{equation}
  \mathbb{E}\bigl\|g_{i,t}\bigr\|^2 \le G^2,
  \quad \forall\,i,t.
\end{equation}
\end{assumption}

\begin{assumption}[Finite QoS Factors]
\label{asm:qos}
Teleportation fidelity, latency, and instability satisfy
\begin{equation}
  0 < F_{i,t} \le 1,\qquad
  0 < \tau_{i,t} \le \tau_{\max},\qquad
  0 \le \sigma_{i,t}^2 \le s_{\max},
\end{equation}
for all $i,t$. Thus the QoS factor
\begin{equation}
  \kappa_{i,t}(\alpha)
  =
  \frac{F_{i,t}^{\alpha}}
       {(\tau_{i,t} + \varepsilon)^{\gamma}(\sigma_{i,t}^2 + \varepsilon)^{\delta}}
\end{equation}
is finite and uniformly bounded below:
\begin{equation}
  \kappa_{i,t}(\alpha) \;\ge\; \kappa_{\min} > 0.
\end{equation}
Hence the trust weights
\begin{equation}
  w_{i,t}(\alpha)
  =
  \frac{p_i \,\kappa_{i,t}(\alpha)}
       {\sum_{j=1}^K p_j \,\kappa_{j,t}(\alpha)}
\end{equation}
are positive and satisfy
\begin{equation}
  0 < w_{\min} \;\le\; w_{i,t}(\alpha) \;\le\; w_{\max} < 1.
\end{equation}
\end{assumption}

\begin{assumption}[Client Independence]
\label{asm:independence}
Conditioned on $\theta_t$, the random variables $g_{i,t}$ and
$(F_{i,t}, \tau_{i,t}, \sigma_{i,t}^2)$ are independent across clients $i$.
\end{assumption}

We further define the geometric dispersion
\begin{equation}
  \rho_t = \max_{i} \bigl\|\theta_{i,t} - \theta_t\bigr\|,
  \qquad
  \rho = \sup_{t} \rho_t,
\end{equation}
which quantifies client drift from the global iterate.

\subsection{A2G Update in Euclidean Space}

In $\mathbb{R}^d$, the general manifold aggregation term reduces to
\begin{equation}
  \Psi(\theta_t,\{\theta_{i,t}\})
  =
  \sum_{i=1}^K w_{i,t}(\alpha) \bigl(\theta_{i,t} - \theta_t\bigr),
\end{equation}
yielding the Euclidean A2G update
\begin{equation}
\label{eq:a2g-euclidean-update}
  \theta_{t+1}
  =
  \theta_t
  - \eta \sum_{i=1}^K w_{i,t}(\alpha)\, g_{i,t}
  + \beta \sum_{i=1}^K w_{i,t}(\alpha)\,(\theta_{i,t} - \theta_t).
\end{equation}
The first term corresponds to stochastic gradient descent with QoS-weighted
gradients; the second term is a geometry-consistent correction toward the
QoS-weighted local models.

\subsection{Single-Step Descent}

\begin{lemma}[Single-Step Descent]
\label{lem:descent}
Under Assumptions~\ref{asm:smooth}--\ref{asm:independence} and a constant
server step size $\eta \le 1/(2L)$, there exist constants
$C_1, C_2, C_3 > 0$ such that
\begin{equation}
\label{eq:descent-lemma}
\begin{aligned}
  \mathbb{E}\bigl[f(\theta_{t+1}) \mid \theta_t\bigr]
  &\;\le\;
  f(\theta_t)
  - C_1 \eta\,\kappa_{\min} \bigl\|\nabla f(\theta_t)\bigr\|^2 \\
  &\quad
  + C_2 \eta^2 G^2
  + C_3 \beta \rho^3.
\end{aligned}
\end{equation}
\end{lemma}

\noindent\emph{Proof Sketch.}
Using $L$-smoothness,
\begin{equation}
  f(\theta_{t+1})
  \;\le\;
  f(\theta_t)
  +
  \bigl\langle \nabla f(\theta_t),\,\theta_{t+1}-\theta_t\bigr\rangle
  +
  \frac{L}{2}\bigl\|\theta_{t+1}-\theta_t\bigr\|^2.
\end{equation}
Substitute the A2G update~\eqref{eq:a2g-euclidean-update} and take the
conditional expectation. The QoS-weighted gradient term is controlled by
Assumptions~\ref{asm:unbiased}--\ref{asm:secondmoment}. Independence
(Assumption~\ref{asm:independence}) and bounded QoS
(Assumption~\ref{asm:qos}) control the geometry term. The deviation term
scales as $\mathcal{O}(\rho^3)$. Choosing $\eta \le 1/(2L)$ yields
\eqref{eq:descent-lemma}. \hfill$\Box$

\subsection{Convergence Rate}

\begin{theorem}[Convergence of A2G]
\label{thm:main}
Under Assumptions~\ref{asm:smooth}--\ref{asm:independence} and constant
$\eta \le 1/(2L)$, the A2G iteration satisfies
\begin{equation}
\label{eq:main-rate}
  \frac{1}{T}
  \sum_{t=0}^{T-1}
  \mathbb{E}\bigl\|\nabla f(\theta_t)\bigr\|^2
  \;\le\;
  \mathcal{O}\!\left(\frac{1}{T\,\kappa_{\min}}\right)
  +
  \mathcal{O}(\eta G^2)
  +
  \mathcal{O}(\beta \rho^3).
\end{equation}
\end{theorem}

\noindent\emph{Proof Sketch.}
Summing Lemma~\ref{lem:descent} over $t=0,\dots,T-1$ and using boundedness of
$f$ from below by $f^\star$ yields
\begin{equation}
\label{eq:a2g-summed-descent}
\begin{aligned}
  C_1 \eta\,\kappa_{\min}
  \sum_{t=0}^{T-1} \mathbb{E}\bigl\|\nabla f(\theta_t)\bigr\|^2
  &\;\le\;
  f(\theta_0) - f^\star \\
  &\quad
  + C_2 T \eta^2 G^2
  + C_3 T \beta \rho^3.
\end{aligned}
\end{equation}

Divide by $T \eta\,\kappa_{\min}$ and absorb constants to obtain
\eqref{eq:main-rate}. \hfill$\Box$

\subsection{Special Cases}

The convergence bound \eqref{eq:main-rate} recovers several important
aggregation rules:

\paragraph*{FedAvg ($\alpha = \beta = 0$).}
In this case $\kappa_{\min} = 1$ and the geometry term vanishes, yielding
\begin{equation}
  \frac{1}{T}
  \sum_{t=0}^{T-1}
  \mathbb{E}\bigl\|\nabla f(\theta_t)\bigr\|^2
  \;\le\;
  \mathcal{O}\!\left(\frac{1}{T}\right)
  +
  \mathcal{O}(\eta G^2).
\end{equation}

\paragraph*{QoS--FedAvg ($\alpha > 0,\,\beta = 0$).}
The rate is accelerated by the QoS minimum factor $\kappa_{\min}$, leading to
\begin{equation}
  \frac{1}{T}
  \sum_{t=0}^{T-1}
  \mathbb{E}\bigl\|\nabla f(\theta_t)\bigr\|^2
  \;\le\;
  \mathcal{O}\!\left(\frac{1}{T\,\kappa_{\min}}\right)
  +
  \mathcal{O}(\eta G^2).
\end{equation}

\paragraph*{Geometry-Only FL ($\alpha = 0,\,\beta > 0$).}
The gradient term reduces to standard FedAvg, while the rate is dominated by
the curvature term $\mathcal{O}(\beta \rho^3)$:
\begin{equation}
  \frac{1}{T}
  \sum_{t=0}^{T-1}
  \mathbb{E}\bigl\|\nabla f(\theta_t)\bigr\|^2
  \;\le\;
  \mathcal{O}\!\left(\frac{1}{T}\right)
  +
  \mathcal{O}(\eta G^2)
  +
  \mathcal{O}(\beta \rho^3).
\end{equation}

\paragraph*{Full A2G ($\alpha > 0,\,\beta > 0$).}
Both QoS acceleration and geometric correction appear explicitly, as in
\eqref{eq:main-rate}; small $\beta$ and bounded dispersion $\rho$ ensure that
the curvature term remains controlled.

\subsection{Riemannian Extension}

\begin{remark}[Riemannian Case]
If $\mathcal{M}$ is a complete Riemannian manifold with bounded curvature and
injectivity radius, the above analysis extends by replacing Euclidean
distances, gradients, and deviations with their Riemannian counterparts. The
qualitative structure of Theorem~\ref{thm:main} remains unchanged, with
$\rho$ interpreted as the maximal geodesic deviation.
\end{remark}

\section{System Setup} 
 
All experiments utilized a single NVIDIA Tesla T4 graphics processing unit (GPU) and CUDA version 12.4 in a high-RAM runtime environment. We use a genome binary classification dataset, partitioned across $N{=}5$ clients with shard size $\approx 100$ samples, following approximately non-IID split. Each client trains a SamplerQNN variational classifier, where circuit angles are treated as circular parameters on $\mathbb{T}^d$ and linear offsets remain in $\mathbb{R}^d$.

\subsection{Data and Pre-processing}

We evaluate A2G on two binary medical imaging datasets. The first is the Breast Cancer Wisconsin dataset from \texttt{scikit-learn}, where each sample is represented by 30 real-valued features extracted from digitized fine-needle aspirate (FNA) images. All features are standardized using \texttt{StandardScaler}, and we apply principal component analysis (PCA) to reduce the input dimension to \(d = 4\) for the quantum model. The second dataset is \emph{Breast-Lesions-USG~| A Curated Benchmark Dataset for Ultrasound Based Breast Lesion Analysis} from The Cancer Imaging Archive (TCIA). From this collection we use patient age and categorical ultrasound descriptors (shape, echogenicity, posterior features, calcifications), discard entries marked as ``not applicable'' or ``indefinable'', and encode the diagnosis as a binary label (benign vs.\ malignant). Categorical attributes are label-encoded, age is cast to numeric, and the resulting features are standardized and reduced via PCA to \(d = 2\). For federated simulation, the sklearn breast-cancer data are partitioned approximately IID across clients to provide a baseline, whereas the Breast-Lesions-USG data are split into non-IID, label-skewed client partitions: each client receives a biased subset of the labels with a guaranteed minimum shard size, emulating heterogeneous and unbalanced clinical sites.

\section{Experimental Settings}

In this section we outline the experimental configurations used to evaluate A2G under varying data heterogeneity, geometric tuning, quantum noise, and scale.

\subsection{Non-IID Federated Partitions}
To emulate realistic client heterogeneity on the Breast-Lesions-USG dataset,
we construct non-IID training partitions and compare them against an
approximately IID baseline.  
In the IID baseline, the shuffled training set is split into $K$ shards of
similar size, each assigned to one client, so that local class histograms
closely match the global distribution.

\paragraph*{Label-skewed setting.}
In the label-skewed configuration, the global training set is grouped by
diagnostic label and then shuffled within each group.  Each client $k$
receives data drawn from a biased subset of labels, so that some classes are
over–represented while others are under–represented or even absent locally.
This produces heterogeneous class mixtures across clients while preserving the
overall global label histogram.

\paragraph*{Quantity-skewed setting.}
In the quantity-skewed configuration, all clients draw from the same global
label distribution, but with unbalanced dataset sizes.  After shuffling the
training indices, we assign to client $k$ a shard whose size is sampled
uniformly from a prescribed interval (e.g., $[20,150]$ examples).  As a
result, some clients act as ``data-rich’’ nodes and others as ``data-poor’’
nodes, while the global class proportions remain unchanged.

These two regimes, together with the IID baseline, allow us to probe the
behaviour of A2G across a spectrum of practically relevant non-IID
configurations.  Fig.~\ref{fig:noniid_a} reports the corresponding
global test accuracy trajectories.

\subsection{Geometry Gain Tuning}

\begin{figure}[!h]
  \centering
  \subfloat[\label{fig:acc1}]{%
  \includegraphics[width=0.48\columnwidth]{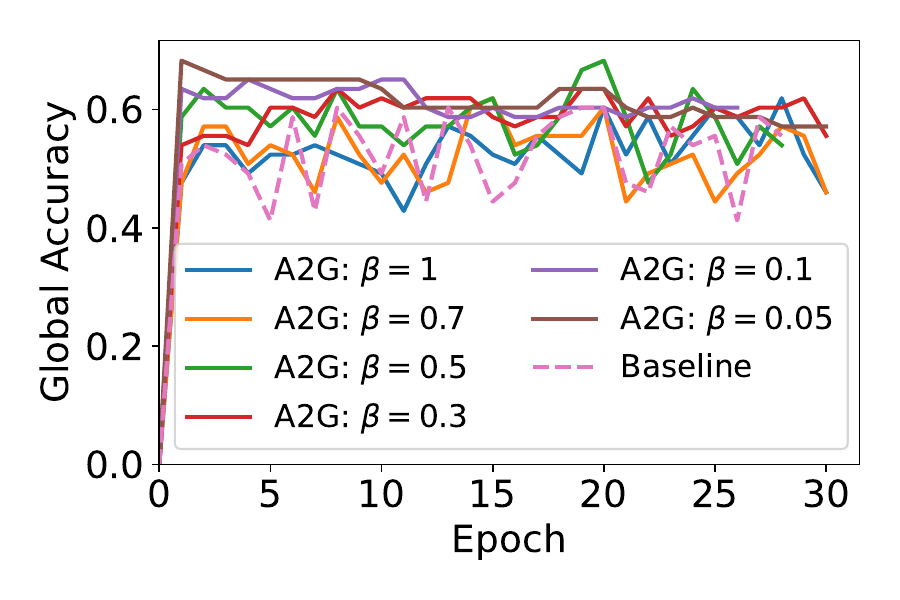}%
  }\hfil
  \subfloat[\label{fig:valloss1}]{%
    \includegraphics[width=0.48\columnwidth]{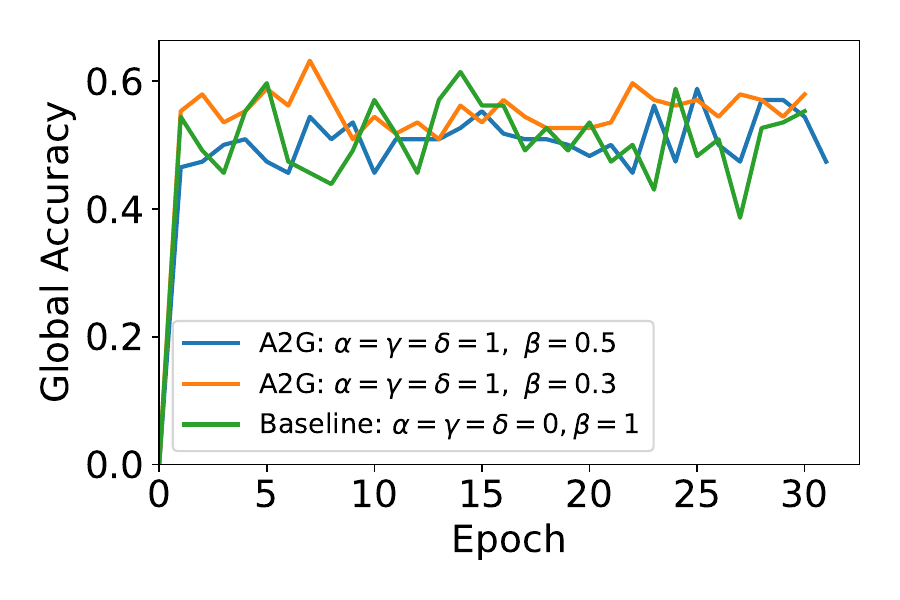}%
  }
  \caption{Geometry gain tuning for A2G on two datasets.
  (a) Breast-Lesions-USG dataset with a non-IID quantity–skew
  partition: global test accuracy vs.\ epoch for the FedAvg baseline
  and A2G with different geometry gains
  $\beta \in \{0.05, 0.10, 0.30, 0.50, 0.70, 1.0\}$ (with
  $\alpha=\gamma=\delta=0$).
  (b) Sklearn Breast Cancer dataset: comparison of QoS-aware A2G
  with $\alpha=\gamma=\delta=1$ and geometry gains
  $\beta \in \{0.3, 0.5\}$ against the FedAvg baseline
  ($\alpha=\gamma=\delta=0,\ \beta=1$).}
  \label{fig:beta_tuning}
\end{figure}

Figs.~\ref{fig:beta_tuning}(a)–(b) and Table~\ref{tab:lesions_beta_sweep}
jointly highlight a non-trivial trade–off induced by the geometry gain~$\beta$.
On the Breast-Lesions-USG dataset, very small geometric corrections
($\beta = 0.05$) already yield a noticeable improvement over the FedAvg
baseline, achieving the highest best accuracy (68.25\%) and the strongest
final and late-epoch performance. As $\beta$ increases beyond $0.1$, the
curvature term in the A2G update becomes dominant, leading to more volatile
learning dynamics and a systematic degradation in both final and mean
accuracy (e.g., $\beta \in \{0.30,0.70,1.00\}$ in
Table~\ref{tab:lesions_beta_sweep}). A similar pattern is observed on the
Sklearn Breast Cancer dataset in Fig.~\ref{fig:beta_tuning}(b): moderate
geometry gains ($\beta \in \{0.3,0.5\}$) consistently outperform the
FedAvg-style baseline, but excessively large corrections fail to deliver
additional benefit. Overall, these results indicate that A2G gains are
maximised in a regime where geometry provides a gentle correction around the
QoS-weighted aggregate, rather than aggressively pulling the global model
towards potentially noisy local optima.
\begin{figure}[!h]
  \centering
  \subfloat[\label{fig:inst1}]{%
  \includegraphics[width=0.48\columnwidth]{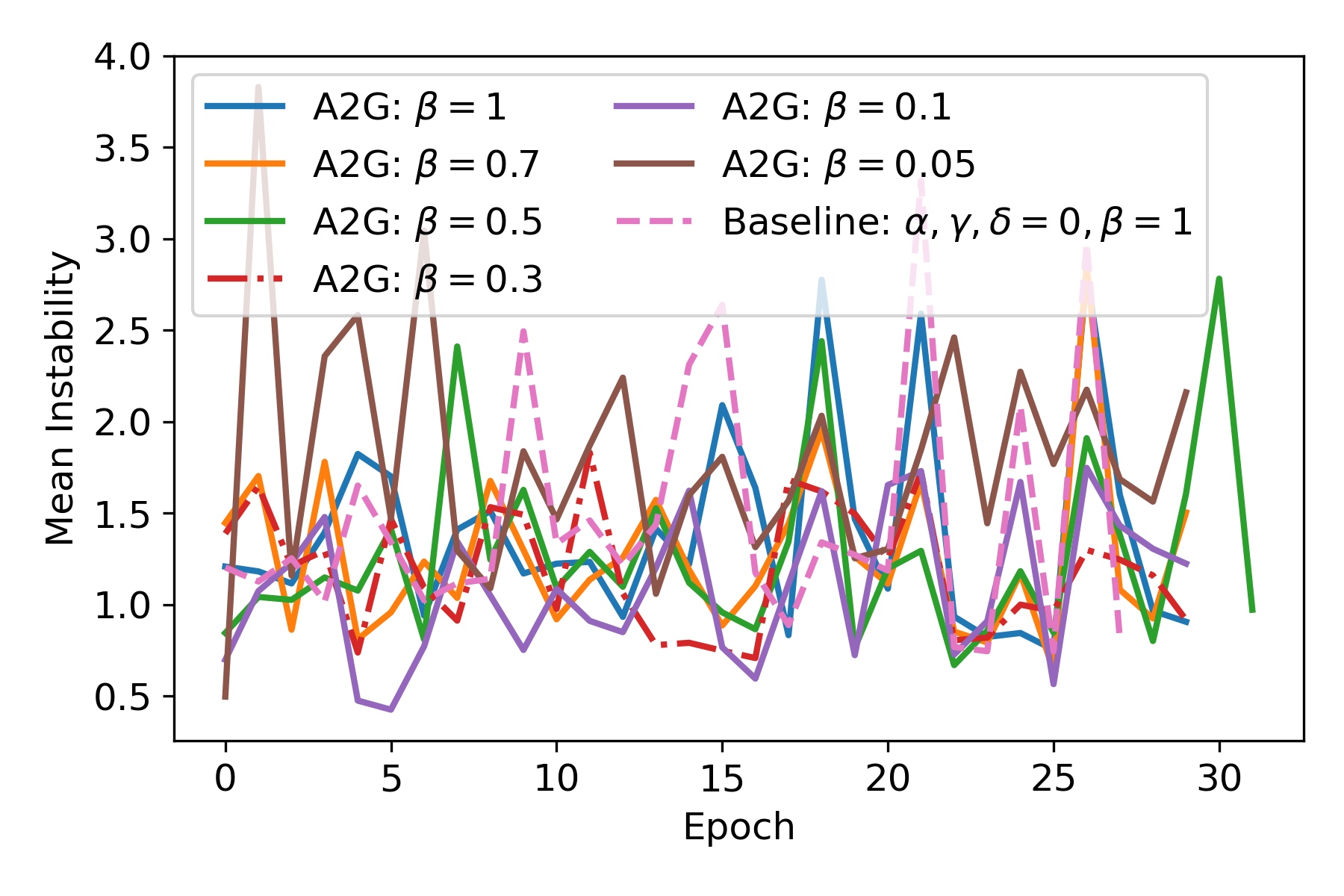}%
  }\hfil
  \subfloat[\label{fig:latency}]{%
    \includegraphics[width=0.48\columnwidth]{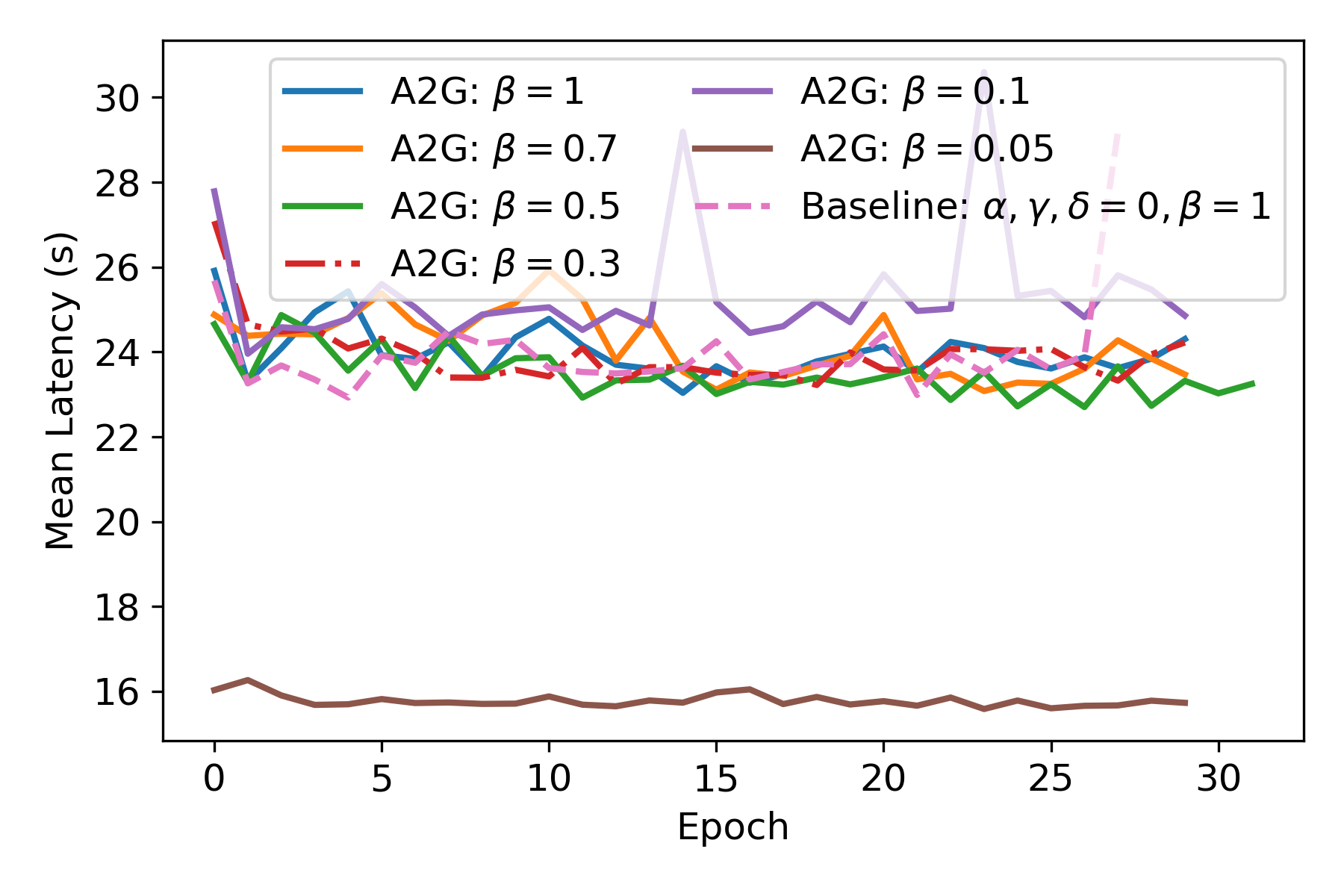}%
  }
  \caption{QoS trends under geometry–gain tuning on the Breast-Lesions-USG
  dataset.
  (a) Mean instability vs.\ epoch for A2G with geometry gains
  $\beta \in \{1.0,0.7,0.5,0.3,0.1,0.05\}$ and the FedAvg baseline
  ($\alpha=\gamma=\delta=0,\ \beta=1$). 
  (b) Mean latency vs.\ epoch for the same runs.  small geometry gains
  (especially $\beta=0.05$) yield consistently lower instability and latency
  than both the baseline and the aggressive geometry setting $\beta=1$.}
  \label{fig:qos_trends}
  \label{fig:6}
\end{figure}

Fig.~\ref{fig:qos_trends} illustrates how the geometry gain $\beta$ shapes
the QoS profile of A2G Fig. \ref{fig:inst1} shows that large geometry gains
($\beta \ge 0.5$) lead to noticeably more volatile instability across
communication rounds, whereas small gains ($\beta \le 0.1$) keep the
QoS-induced variability tightly bounded.  Fig. \ref{fig:latency} reports the mean latency
per round: here, the conservative setting $\beta=0.05$ achieves the lowest and
most stable latency, while the FedAvg baseline and $\beta=1$ incur both higher
average delay and larger spikes.  Together with the accuracy results, these
trends indicate that moderate geometric updates not only improve test
performance but also stabilise the communication behaviour of the federated
system under teleportation noise.

\begin{table}[t]
\centering
\caption{Geometry gain sweep on Breast-Lesions-USG (quantity-skew,
medium teleportation noise). Global accuracy is reported in percent.}
\label{tab:lesions_beta_sweep}
\begin{tabular}{ccccc}
\toprule
$\beta$ 
& \#Epochs 
& Best acc. 
& Final acc. 
& Mean acc. (last 5) \\
\midrule
0.05 & 20 & 68.25 & 63.49 & 62.22 \\
0.10 & 26 & 65.08 & 60.32 & 60.63 \\
0.30 & 30 & 63.49 & 55.56 & 59.37 \\
0.70 & 30 & 61.90 & 46.03 & 52.06 \\
1.00 & 30 & 61.90 & 46.03 & 54.60 \\
\bottomrule
\end{tabular}
\end{table}

\subsection{Teleportation Noise Regimes}

\begin{figure}[!h]
  \centering
  \subfloat[\label{fig:noniid_a}]{%
  \includegraphics[width=0.48\columnwidth]{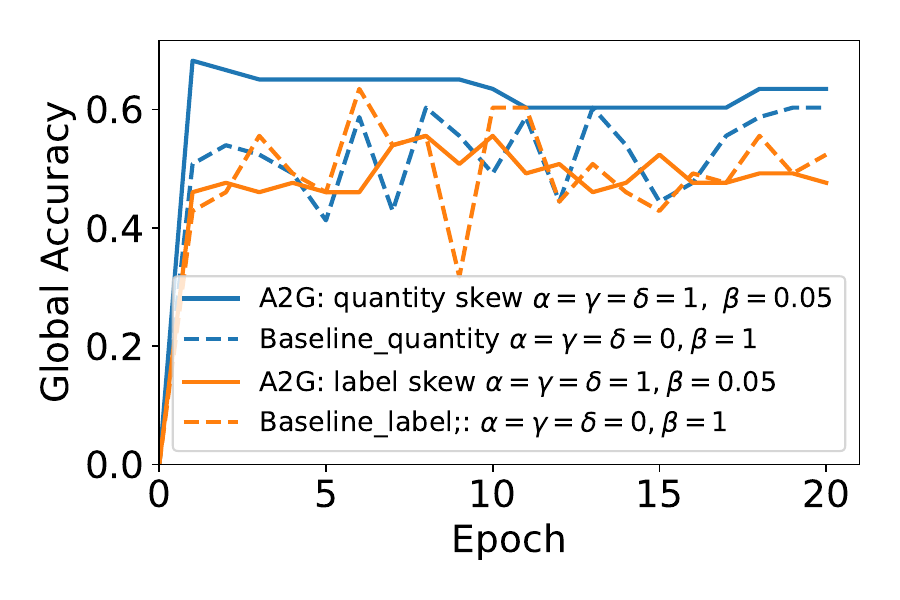}%
  }\hfil
  \subfloat[\label{fig:noniid_b}]{%
    \includegraphics[width=0.48\columnwidth]{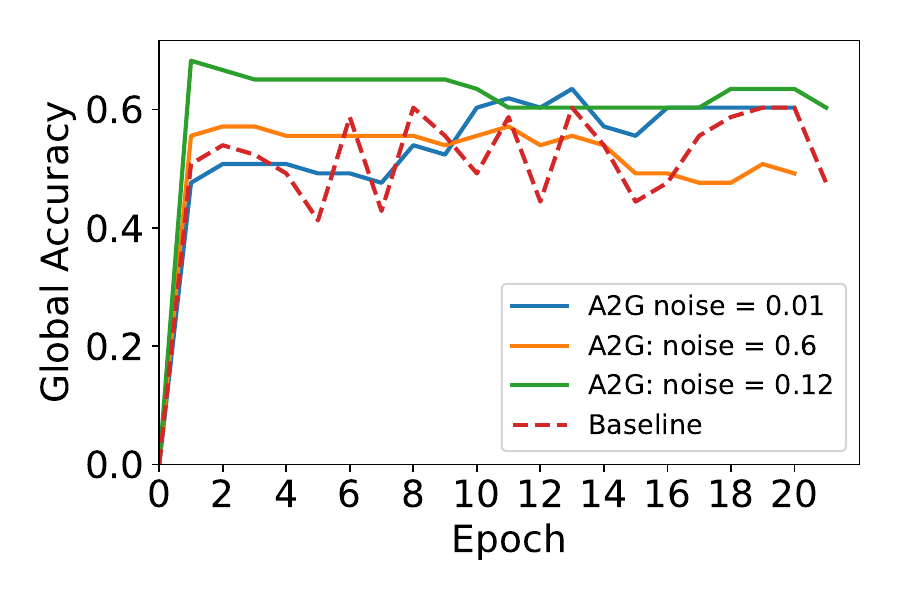}%
  }
  \caption{Impact of A2G under non-IID partitions
  and different noise regimes on the Breast-Lesions-USG dataset.
  (a) Global test accuracy vs.\ epoch for quantity-skewed and
  label-skewed client partitions, comparing A2G with
  $\alpha=\gamma=\delta=1,\ \beta=0.05$ against the FedAvg baseline
  ($\alpha=\gamma=\delta=0,\ \beta=1$) under a medium-noise
  teleportation channel ($p=0.06$).
  (b) Global test accuracy vs.\ epoch for different teleportation noise
  levels $p \in \{0.01, 0.06, 0.12\}$ using A2G
  ($\alpha=\gamma=\delta=1,\ \beta=0.05$), contrasted with the FedAvg
  baseline without teleportation.}
  \label{fig:noise_nonIID}
\end{figure}

To study the robustness of A2G under degraded quantum communication,
we model the teleportation link as a synthetic bit-flip channel with
configurable error probability $p$.  Each teleportation trial incurs an
independent bit flip with probability $p$, which lowers the empirical
fidelity samples $F_{i,t}$ used in the QoS module.  We consider three
noise regimes $ p \in \{0.01,\, 0.06,\, 0.12\},$
corresponding to \emph{low}, \emph{medium}, and \emph{high} noise,
respectively.  

Fixing the geometry gain to $\beta = 0.05$ and the QoS exponents to
$\alpha = \gamma = \delta = 1$, we rerun the federated training on the
Breast-Lesions-USG dataset for each noise level and non-IID partition
setting.  For every run we log global test accuracy, latency statistics
(mean and 90th percentile), and the QoS-derived trust weights, and we
contrast A2G with a FedAvg baseline that does not employ teleportation
($p = 0$).  This enables us to characterise how A2G trades off
performance and robustness as the effective fidelity of the
teleportation link deteriorates (Fig.~\ref{fig:noniid_b}).

\section{Discussion}
In this work we deliberately adopt a geometry–gain update rather than a gradient-based
server update. This choice is principled for both classical and quantum FL.
In variational quantum circuits, server-side gradients are notoriously expensive and
fragile due to sampling noise and barren-plateau effects~\cite{
cerezo2021variational}, and parameter-shift evaluations incur a substantial communication
and runtime overhead when aggregated across many clients~\cite{schuld2019quantum}.
These issues are exacerbated when model updates must be transmitted over noisy
or teleportation-based links in QFL~\cite{li2020federated}.
By contrast, client-side optimisers such as SPSA can estimate local gradients with
low communication cost and are already widely used in noisy quantum settings.
Restricting the server to a geometry-controlled interpolation step therefore avoids
expensive global gradient computation while still steering the model towards
consensus.

The geometry gain parameter $\beta$ provides a stable, communication-efficient,
gradient-free mechanism to blend the previous global model with the QoS-weighted
aggregate. Small but nonzero values (e.g., $\beta=0.05$ on Breast-Lesions-USG)
consistently outperform the FedAvg-style setting ($\beta=1.0$) by more than
ten percentage points in best and late-round accuracy, indicating that
geometry-aware damping can mitigate both client drift and QoS-induced bias.
This behaviour is consistent with observations from classical FL, where
momentum, EMA-style smoothing, and proximal or control-variate corrections
are used to stabilise aggregation under heterogeneity~\cite{mcmahan2017communication,
karimireddy2020scaffold,li2020federated}. Our results therefore suggest that
geometry-gain A2G offers a practically attractive and theoretically grounded
alternative to gradient-based server updates in both quantum and hybrid
quantum--classical federated deployments.

\section{Conclusion}

We introduced A2G-QFL (Adaptive Aggregation with QoS and Geometry Gains), 
a unified aggregation framework designed for heterogeneous, latency-sensitive, and 
quantum-enabled federated learning systems. A2G departs from conventional federated 
optimization by integrating two complementary mechanisms: 
(i) \emph{teleportation-aware QoS gains}, which transform fidelity, latency, and 
instability indicators into principled trust weights; and 
(ii) \emph{geometry gains}, which introduce a curvature-consistent correction that 
reduces client drift and improves robustness under heterogeneous local updates. 
These components produce an aggregation rule that strictly generalises FedAvg, QoS-FL, 
and Riemannian FL, while naturally supporting quantum, classical, and hybrid model architectures.
Overall, A2G offers a \emph{general, theoretically grounded, and communication-aware} 
aggregation strategy that strengthens convergence guarantees under both classical and 
quantum communication constraints. Its modular structure, compatibility with arbitrary 
optimizers and model classes, and principled incorporation of QoS and geometry make A2G 
a promising foundation for next-generation distributed and quantum-enabled learning frameworks.

\bibliographystyle{IEEEtran}
\bibliography{refs}

\end{document}